%% file: main.tex
\documentclass[10pt,twocolumn,letterpaper]{article}

\usepackage[pagenumbers]{cvpr} 


\definecolor{cvprblue}{rgb}{0.21,0.49,0.74}
\usepackage[pagebackref,breaklinks,colorlinks,allcolors=cvprblue]{hyperref}
\usepackage{amsfonts}
\usepackage{amsmath}
\usepackage{bm}
\usepackage{booktabs}
\usepackage{multirow} 
\usepackage{float}

\newcommand{\eref}[1]{Eq. \eqref{#1}}
\newcommand{\tref}[1]{Table \ref{#1}}
\newcommand{\fref}[1]{Figure \ref{#1}}
\newcommand{\aref}[1]{Appendix \ref{#1}}

\def \x {\bm{x}}
\def \y {\bm{y}} 
\def \p {\bm{p}}
\def \P {\textbf{P}}
\def \v {\bm{v}}
\def \t {\bm{t}}
\def \g {\bm{g}}
\def \z {\bm{z}}
\def \l {\bm{l}}
\def \q {\bm{q}}
\def \S {\bm{S}}

\def\paperID{18252} 
\def\confName{CVPR}
\def\confYear{2025}

\title{Context-Based Semantic-Aware Alignment for \\Semi-Supervised Multi-Label Learning}

\author{
Heng-Bo Fan$^{1}$\thanks{Equal contributions.} \quad 
Ming-Kun Xie$^{1}$\footnotemark[1] \quad 
Jia-Hao Xiao$^{1}$ \quad 
Sheng-Jun Huang$^{1}$\thanks{Corresponding author.}  \\
$^1$Nanjing University of Aeronautics and Astronautics, Nanjing, China \\
{\tt\small\{fan\_heng\_bo, mkxie, jiahaoxiao, huangsj\}@nuaa.edu.cn} \\
}

\begin{document}
\maketitle
\input{sec/0_abstract}    
\input{sec/1_intro}
\input{sec/2_related_work}
\input{sec/3_method}
\input{sec/4_experiment}
{
    \small
    \bibliographystyle{ieeenat_fullname}
    \bibliography{main}
}

\appendix
\input{sec/X_suppl}

\end{document}

%% file: sec/0_abstract.tex
\begin{abstract}
Due to the lack of extensive precisely-annotated multi-label data in real word, semi-supervised multi-label learning (SSMLL) has gradually gained attention. Abundant knowledge embedded in vision-language models (VLMs) pre-trained on large-scale image-text pairs could alleviate the challenge of limited labeled data under SSMLL setting.
Despite existing methods based on fine-tuning VLMs have achieved advances in weakly-supervised multi-label learning, they failed to fully leverage the information from labeled data to enhance the learning of unlabeled data. In this paper, we propose a context-based semantic-aware alignment method to solve the SSMLL problem by leveraging the knowledge of VLMs. To address the challenge of handling multiple semantics within an image, we introduce a novel framework design to extract label-specific image features. This design allows us to achieve a more compact alignment between text features and label-specific image features, leading the model to generate high-quality pseudo-labels. To incorporate the model with comprehensive understanding of image, we design a semi-supervised context identification auxiliary task to enhance the feature representation by capturing co-occurrence information. Extensive experiments on multiple benchmark datasets demonstrate the effectiveness of our proposed method. 
\end{abstract}

%% file: sec/1_intro.tex
\section{Introduction}
\label{sec:intro}

In contrast to traditional multi-class classification, which assumes that each instance belongs to a single class, multi-label learning (MLL) aligns more closely with real-world scenarios where each image may contain multiple semantic objects simultaneously. This complexity arises from diverse shapes and sizes of objects contained in an image, making MLL a particularly challenging task.
Thanks to the powerful capacity of deep neural networks (DNNs), MLL has achieved significant success in various domains, \textit{e.g.}, video annotation \cite{kang2006correlated}, the recommendation system \cite{manoharan2022optimized, carrillo2013multi}, and medical diagnosis \cite{ge2020improving, chen2020deep}.

\begin{figure}[t]
  \centering
   \includegraphics[width=1.0\linewidth]{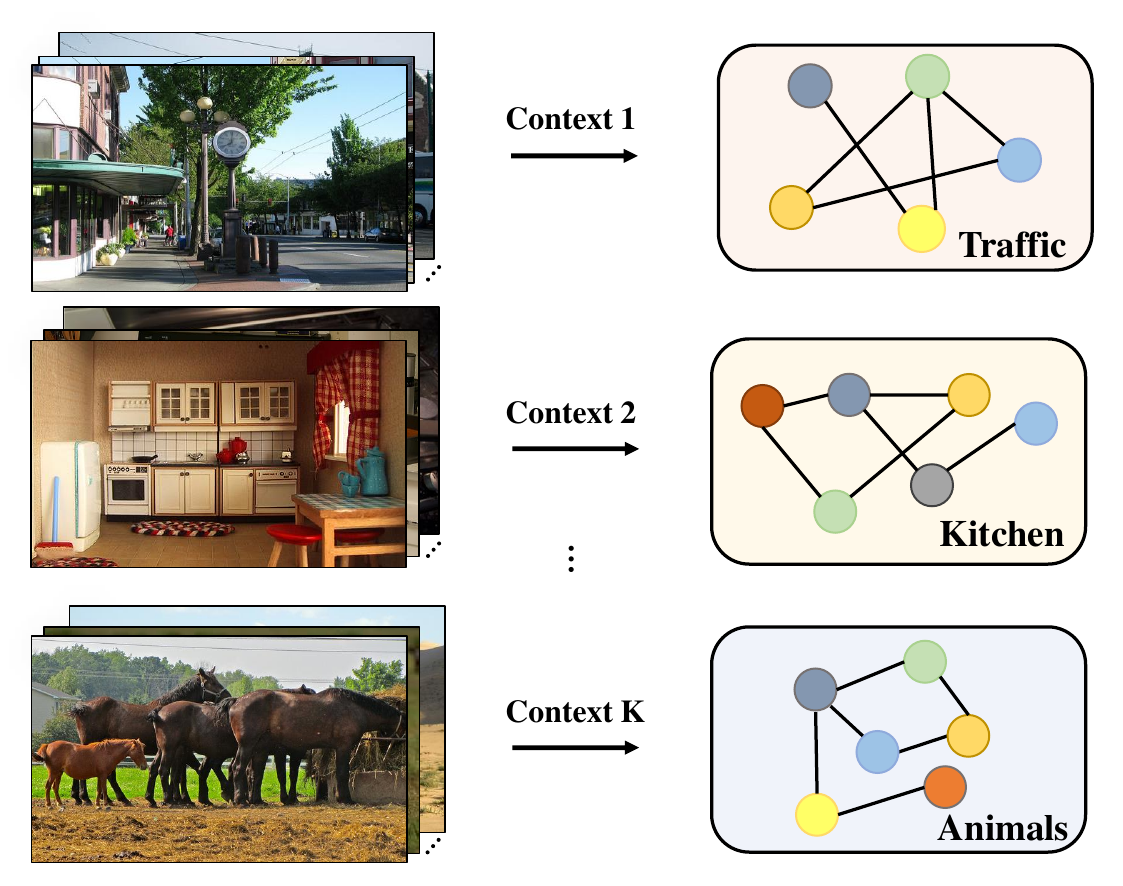}
   \caption{In MLL, context information determines the co-occurrence relationships among categories. Semantics belonging to the same context are more likely to appear simultaneously. To leverage this, we introduce an auxiliary task: Context Identification, during model training. This task helps narrow down the label space into its relevant subset.}
   \label{fig:1stdemo}
\end{figure}

While multi-label learning (MLL) has recently made significant strides, a potential challenge lies in the fact that training an effective DNN often demands extensive, accurately-annotated multi-label data, which can be difficult and costly to collect in real world. Semi-supervised multi-label learning (SSMLL) provides a solution to this problem by training a MLL model based on a limited set of labeled examples with a substantial amount of unlabeled examples.

Recently, vision-language models (VLMs) have made significant progress in various fields, benefiting from their pre-training on large-scale datasets consisting of image-text pairs. This motivates us to leverage the abundant knowledge embedded in VLMs to alleviate the challenge of limited labeled data. Several efforts have been made to incorporate VLMs into multi-label learning with partial labeling \cite{sun2022dualcoop, ding2023exploring}, yielding promising results in this task. 
Unfortunately, since these methods are not specifically designed for SSMLL, they cannot fully leverage the precisely-labeled examples to enhance the learning process on unlabeled data, resulting in sub-optimal performance. To the best of our knowledge, in this paper, we make the first attempt to fine-tune VLMs in SSMLL downstream task, which aims to leverage the information of unlabeled data and the knowledge embedded in VLMs simultaneously. 

One of the primary challenges in Multi-Label Learning (MLL) lies in exploring the correlations among labels. Existing methods often capture label correlations by either adopting graph convolutional network \cite{chen2019multi} or developing label masking strategy \cite{xu2022boosting}. In practice, we observe that whether a target is present in an image often depends on the context of the image. For example, as shown in \fref{fig:1stdemo}, an image captured in a street scene is more likely to contain objects such as \textit{car}, \textit{people}, or \textit{traffic light}; whereas in an image taken in a kitchen, it is more probable to have objects like \textit{refrigerator}, \textit{oven}, or \textit{dining table}. In other words, the co-occurrence of these labels constitutes the context. This motivates us to identify image contexts to capture co-occurrence relationships, thereby enhancing the performance of SSMLL.

In this paper, we propose a context-based semantic-aware alignment method to solve the SSMLL problem, which aims to improve pseudo-labeling performance by leveraging the knowledge of CLIP. On the one hand, we develop the semantic-aware alignment target task to encourage a more compact alignment between text features and image features. The core idea involves aligning text features with label-specific features, rather than features of the entire image. This simplifies the alignment task, contributing to a more effective utilization of knowledge embedded in CLIP. On the other hand, we design a context identification auxiliary task to explore co-occurrence information. We first divide the original dataset into multiple clusters, each representing a context category; then, we formulate the context identification task as a semi-supervised learning problem to leverage the co-occurrence relationships. Experimental results on multiple benchmark datasets demonstrate that our method outperforms existing fine-tuning methods.

%% file: sec/2_related_work.tex
\vspace{-0.5em}
\section{Related Work}
\label{sec:formatting}
\subsection{Multi-label Learning}
Multi-label learning has experienced rapid development in recent years. 
And these works can be mainly classified into three groups. The first group attempted to capture the correlation between labels using graph-based \cite{chen2019multi, chen2020deep} methods, RNN/LTSM \cite{wang2017multi, yazici2020orderless}, or transformer \cite{pu2022semantic, yuan2023graph}. Correlation between different labels can be used as prior knowledge for classification task. 
The second group improved loss function \cite{guo2021long, ridnik2021asymmetric, kim2022large, huang2023asymmetric} to alleviate the problem of extreme imbalance ratio between positive and negative labels. The final group designed more sophisticated model structures to better extract local image feature using attention-based \cite{sovrasov2022combining, lanchantin2021general} technique.

Due to limited annotations in real-world scenario, more realistic problem settings have been proposed in MLL. 
For instance, \textit{partial label} \cite{xie2018partial, xie2021partial} setting, where a subset of original labels is available for each training instance. \textit{Single positive label} \cite{cole2021multi} setting is a more extreme case where merely one single positive label is provided for every instance.
However, the field of SSMLL is relatively unvisited. 
Wang \textit{et al.} \cite{wang2013dynamic} and Zhao \textit{et al. } \cite{zhao2015semi} leveraged label relations in non-deep scenario, which cannot be applied to our case. Wang \textit{et al.} \cite{wang2020dual} used two classifier to deal with the features of labeled and unlabeled samples separately, which could reduce the distribution gap between these two datasets. However, due to the extreme limited supervised information in SSMLL, previous methods failed to generate accurate pseudo-labels. 
Thus, in this paper, we leverage VLMs pre-trained on numerous image-text pairs to alleviate the challenge of limited labeled data.


\begin{figure*}[t!]
  \centering
   \includegraphics[width=1.0\linewidth]{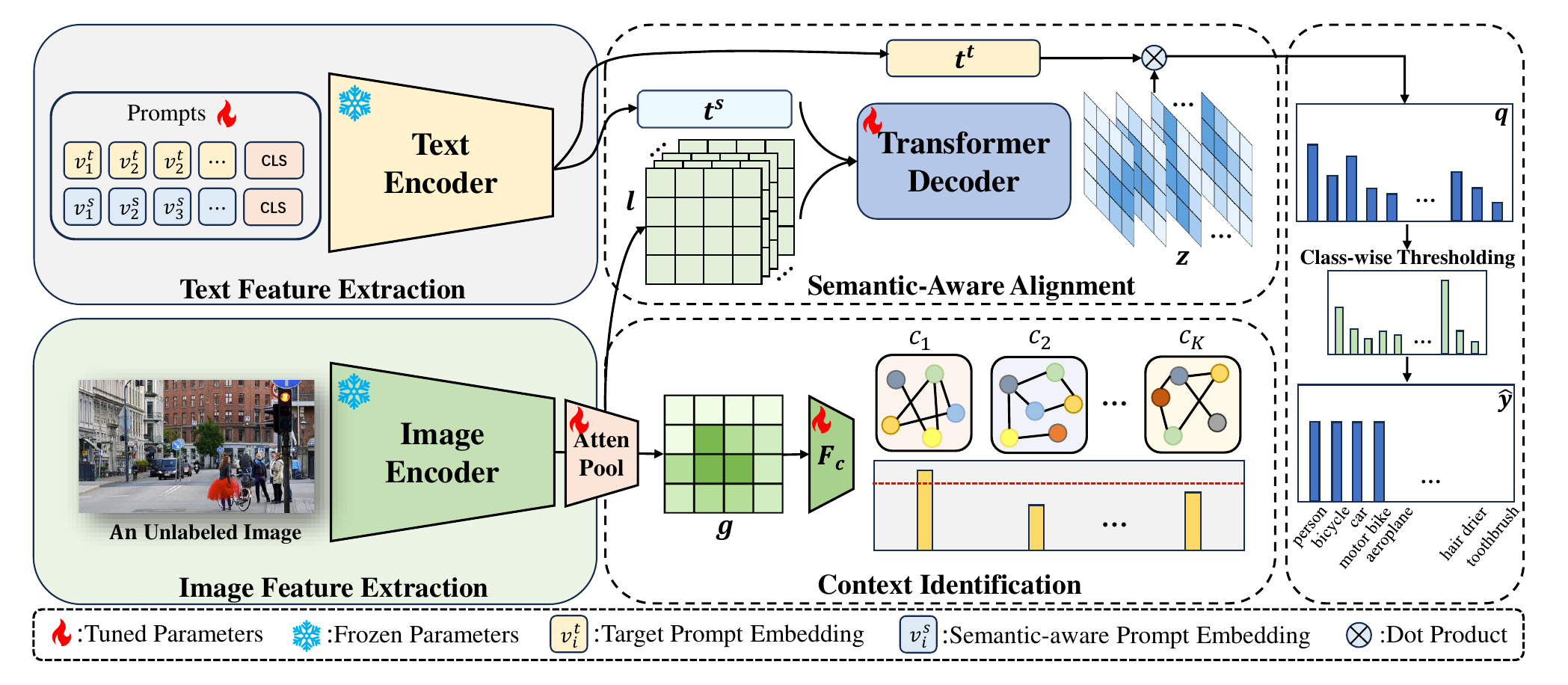}

   \caption{An overview of our proposed method. We begin by employing CLIP to extract both image features and $C$-class text features. Subsequently, we thoroughly leverage the pre-training knowledge embedded in the text encoder of CLIP to extract label-specific image features. To further exploit label correlation, we introduce a \textbf{Context Identification} task. Finally, we optimize the network by generating pseudo-labels with class-wise thresholds.}
   \label{fig:2nddemo}
\end{figure*}

\subsection{Overview of CLIP}
VLMs have achieved tremendous success in various fields in recent years. As a milestone of VLMs, CLIP, pre-trained with 400 million image-text pairs, consists of an image encoder and a text encoder. The optimizing objective of CLIP is to align visual and textual modalities in latent space using a contrastive loss. Benefited from the tight alignment of two modalities, CLIP has exhibited promising result on classification tasks simply by using prompt as query. One naive way to construct prompt is using a template like "a photo of a [CLS].", where [CLS] can be substituted with real category names. 
\cite{zhou2022learning} has demonstrated that optimizing continuous learnable prompt $\boldsymbol{V}$ rather than using hand-crafted ones could achieve better performance on different tasks.  Given an image input $\boldsymbol{x}_i$, the two modal features can be computed as follows
\begin{equation}
    \boldsymbol{h}_i=\mathtt{Enc}_{\text{I}}(\boldsymbol{x}_i),
\boldsymbol{w}=\mathtt{Enc}_{\text{T}}(\boldsymbol{V}),
\end{equation}
and then CLIP computes cosine similarity between image feature $\boldsymbol{h}_i$ and text feature $\boldsymbol{w}$. Finally we optimize learnable prompts by maximizing model's prediction on ground-truth label
\begin{equation}
    p(y|\boldsymbol{x}_i)=\frac{\exp(\langle\boldsymbol{h}_i, \boldsymbol{w}\rangle/\tau)}{\sum_{j=1}^C\exp(\langle\boldsymbol{h}_i, \boldsymbol{w}_j\rangle/\tau)},
\end{equation}
where $\langle\cdot,\cdot\rangle$ compute cosine similarity and $\tau$ is a temperature parameter.

Follow-up works introduced various prompt designs like adding prompts to visual modality \cite{jia2022visual, ge2022domain, shu2022test, shen2022multitask, zhang2021domain}, or using multiple prompts simultaneously \cite{sun2022dualcoop, guo2023texts}. More adapter designs \cite{zhang2022tip, zhang2023prompt, gao2023clip, rao2022denseclip} have been proposed to further increase CLIP's transfer ability on downstream task. While its powerful generalization ability on partial multi-label learning \cite{sun2022dualcoop} and zero-shot multi-label learning \cite{ding2023exploring} have been validated, adapting CLIP into SSMLL is still unexplored. Furthermore, these approaches failed to consider the unique challenge of multi-label learning, and fine-tuned CLIP by aligning multiple class text embeddings with one single image embedding. This results in sub-optimal model performance due to its complex optimization objective. 
Therefore, we propose to simplify the alignment task and fully leverage contextual information within an image to enhance the quality of pseudo labels. This results in a more compact alignment between visual and textual modalities, providing a solution to SSMLL problem. 

%% file: sec/3_method.tex
\vspace{-1em}
\section{Proposed Method}
In this section, we present Context-Based Semantic-Aware Alignment framework to adapt CLIP to SSMLL. We use $\boldsymbol{x}\in\mathcal{X}$ to denote an image, and $\boldsymbol{y}\in\mathcal{Y}$ to denote its label, where $\mathcal{X}\in\mathbb{R}^d$ represents the feature space and $\mathcal{Y}\in\{0,1\}^C$ represents the label space. Under the SSMLL problem setting, the whole training dataset can be divided into a labeled dataset $\mathcal{D}_l=\{(\boldsymbol{x}_i, \boldsymbol{y}_i)\}_{i=1}^m$ and an unlabeled dataset $\mathcal{D}_u=\{\boldsymbol{x}_j\}_{j=1}^n$. For an image $\boldsymbol{x}_i\in\mathcal{D}_l$, its label vector $\boldsymbol{y}_i$ is a multi-hot vector, where $y_{ik}=1$ indicates the presence of object $k$ in image $\boldsymbol{x}_i$, while $y_{ik}=0$ indicates its absence.

The primary challenge posed by SSMLL lies in how to generate high-quality pseudo-labels for unlabeled instances based on a small number of labeled examples. To address this challenge, we improve the quality of pseudo-labels from two aspects: 1) develop the semantic-aware alignment target task to enforce a compact alignment between two modalities; 2) design the context identification auxiliary task to exploit co-occurrence relationships. \fref{fig:2nddemo} illustrates the whole pipeline of our framework. Below, we will introduce the details of these two components.

\subsection{Semantic-Aware Alignment}

\begin{figure}[t]
  \centering
	\includegraphics[width=0.8\linewidth]{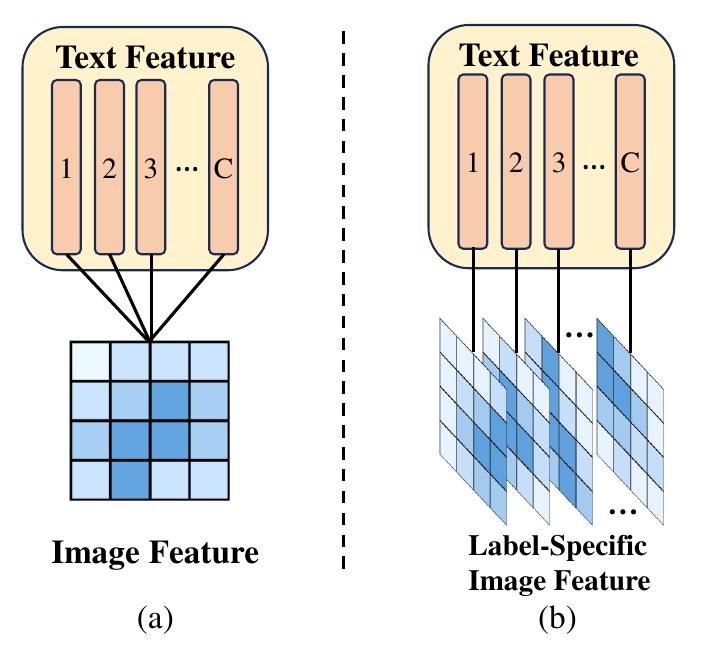}
   \caption{A comparison between previous alignment task and ours. (a) Previous methods align multiple text embeddings with one whole image embedding. Instead, (b) our method extracts label-specific image features and aligns them with the corresponding class text features.}
 \label{fig:label_specific}
 \vspace{-1em}
\end{figure}

Instead of fine-tuning the entire CLIP, the existing methods \cite{zhou2022learning, sun2022dualcoop} often fine-tune the learnable prompts by aligning text features with image features. However, as shown in \fref{fig:label_specific}, when tackling instances associated with multiple labels, it transforms the alignment task from a one-to-one task, \textit{i.e.}, aligning one text embedding with one image embedding, into a many-to-one problem, \textit{i.e.}, aligning multiple text embeddings with one image embedding. This often results in the sub-optimal alignment between text features and image features, leading to unfavorable fine-tuning performance. To address this problem, we propose to perform semantic-aware alignment between the text features and label-specific image features rather than the features of a whole image. This simplifies the alignment task to an one-to-one problem, leading to a better alignment between two modalities.

Specifically, for each unlabeled instance $\x_j$, by feeding it into the image encoder $\text{Enc}_{\text{I}}(\cdot)$, we obtain the image features $\boldsymbol{f}_j=\text{Enc}_{\text{I}}(\x_j)$. Taking ResNet \cite{he2016deep} as an example, CLIP slightly modifies its structure by adding a attention pooling layer before taking the  inner product with text feature. CLIP applies average pooling to $\boldsymbol{f}_j \in\mathbb{R}^{HW\times d}$ to obtain global image feature $\bar{\boldsymbol{f}_j}\in \mathbb{R}^{1\times d}$, where $H$, $W$, $d$ are the height, width and the number of channels. 
Then, the concatenated image features $[\bar{\boldsymbol{f}_j}, \boldsymbol{f}_j]$ is fed into a multi-head self-attention layer to obtain the global image feature $\g_j$ and feature map $\l_j$:
\begin{equation}\label{eq:ap}
	[\g_j, \l_j] = \mathtt{SelfAttention}([\bar{\boldsymbol{f}_j}, \boldsymbol{f}_j]).
\end{equation}

In the original CLIP, only the global image $\g_j$ feature is used to compute the similarity with text feature. However, we find that the feature map $\l_j$ still has abundant semantics and can be used to extract label-specific features. Thus we denote $\l_j$ as local image feature.

Suppose the length of learnable prompt is $N$, so for every class $k$, we use a semantic-aware prompt, $\P_k^s=[\v^s_1,\v^s_2,...,\v^s_N,\text{CLS}_k]$ to allocate attention on regions of interest and a target prompt $\P_k^t=[\v^t_1,\v^t_2,...,\v^t_N,\text{CLS}_k]$ to perform target multi-label alignment, where $\v^s_j$ and $\v^t_j$ are learnable word embeddings and $\text{CLS}_k$ is the word embedding of the $k$-th class name. By feeding the prompts into the text encoder of CLIP $\text{Enc}_{\text{T}}(\cdot)$, we obtain the semantic-aware text embedding $\t^s_k$ and target text embedding $\t^t_k$.
We perform cross attention between $\{\t^s_k\}_{k=1}^C$ and $\l_j$, i.e. $\{\t^s_k\}_{k=1}^C$ as query, $\l_j$ as key and value, to generate label-specific image features:
\begin{equation}\label{eq:ca}
\{\z^{(j)}_{k}|k\in[C]\}=\mathtt{TransFormerDec}(\{\t^s_k\}_{k=1}^C,\l_j,\l_j),
\end{equation}
where $\z^{(j)}_{k}$ is the feature of the $k$-th class for instance $\x_j$. Compared to the image features that encodes the information of a whole image, label-specific features capture the information of each semantic object in an image. 
Then, we align the target text features $\t^t_k$ and label-specific image features $\z_k$ in an one-to-one manner and obtain the alignment degree:
\begin{equation}
\p^t_j=\left[\sigma(\langle\z^{(j)}_1,\t^t_1\rangle),...,\sigma(\langle\z^{(j)}_C,\t^t_C\rangle)\right],
\end{equation}
where $\sigma(\cdot)$ is the sigmoid function. For simplicity, we use $\q^t_j$ and $\p^t_j$ to denote the alignment degrees on the weakly-augmented and strongly-augmented versions of images.

\paragraph{Semi-Supervised Semantic-Aware Alignment.} In order to leverage the information of unlabeled instances, we generate the pseudo-labels using class-distribution-aware thresholding (CAT) \cite{xie2023class}:
\begin{equation}
	\hat{y}_{jk}=\begin{cases}1&\mathrm{if~}q_{jk}\geq\tau^+_k,\\0&\mathrm{if~}q_{jk}\leq\tau^-_k,\\-1&\mathrm{otherwise},\end{cases}
\end{equation}
where $\tau^+_k$ and $\tau^-_k$ are the class-wise thresholds that respectively capture the positive and negative proportions in the unlabeled data.
Given the predicted degrees $\p^t_j$ and pseudo-labels $\hat{\y}_j$, we define the unsupervised alignment loss as follows:
\begin{equation}
\mathcal{L}_{unsup}=\sum_{j=1}^n\sum_{k=1}^C L(p^t_{jk}, \hat{y}_{jk}),
\end{equation}
here, $L(p_{k}, y_{k})=y_kL_{+}(p_k)+(1-y_k)L_{-}(p_k)$ is the ASL loss \cite{ridnik2021asymmetric}, where $L_{+}(p_k)=-(1-p_k)^{\gamma_1}\log(p_k)$ and $L_{-}(p_k)=-(p_k)^{\gamma_2}\log(1-p_k)$. $\gamma_1$ and $\gamma_2$ are focusing parameters to adjust focusing levels of the positive and negative examples. We define the supervised alignment loss as:
\begin{equation}
\mathcal{L}_{sup}=\sum_{i=1}^m\sum_{k=1}^C L(p^t_{ik}, y_{ik}).
\end{equation}

\subsection{Context Identification}
It is fundamental to exploit the label correlations, which has been regarded as an essential information for MLL. Existing methods often capture label correlations by either using graph convolutional network \cite{chen2019multi} or designing label masking strategy \cite{xu2022boosting}. Unlike these works, we propose to design an auxiliary task that identifies the context of an image for capturing co-occurrence information. Specifically, our approach involves partitioning the original label space into several groups, each representing a context category comprising fewer co-occurring semantic classes. Subsequently, we formulate the context identification task as a semi-supervised classification problem to improve the learning of co-occurrence features for both labeled and unlabeled instances, aiming to enhance the performance of pseudo-labeling.
\vspace{-1em}
\paragraph{Self-Supervised Context Partition.} 
To capture the correlations among labels, we partition the original label set into several context subsets based on label co-occurrence in a self-supervised manner. Specifically, we begin by defining a $C\times C$ co-occurrence matrix $\S$, where $S_{kl}=\frac{e_{kl}}{n_k}$. Here, $e_{kl}$ represents the number of images containing both the $k$-th and $l$-th labels, and $n_k$ represents the number of images containing only the $k$-th label. Inspired by \cite{xu2022boosting}, we apply spectral clustering to the co-occurrence matrix $\S$ to partition the label space into several clusters. Here the affinity matrix of category graph can be computed as follows:
\begin{equation}
\boldsymbol{P}=\frac{\boldsymbol{S}+\boldsymbol{S}^T}{2}.
\end{equation}
The context partition is then formulated as a normalized graph-cut problem:
\begin{equation}
	\hat{\bm{F}}\leftarrow\arg\min_{\boldsymbol{F}}\mathrm{Trace}(\boldsymbol{F}^{\top}\boldsymbol{D}^{-\frac12}\boldsymbol{L}\boldsymbol{D}^{-\frac12}\boldsymbol{F}),\mathrm{~s.t.~}\bm{F}^{\top}\bm{F}=\bm{I},
\end{equation}
here, $\boldsymbol{D}$ is a diagonal matrix, where the diagonal element $D_{kk}=\sum_{k=1}^C S_{kl}$, and $\boldsymbol{L}=\boldsymbol{D}-\boldsymbol{P}$ is the Laplacian matrix. $\bm{F}$ represents the learned graph embeddings of classes, which captures the co-occurrence information of each class. Its solution $\hat{\bm{F}}$ is the eigenvectors corresponding to the top-$k$ minimum eigenvalues. By applying $k$-means on $\hat{\bm{F}}$, we divide the original label set into $K$ subsets. For each labeled instance $\x_i$, we determine its context label $c_i$ based on which subset its semantic labels belong to. 

\vspace{-1em}
\paragraph{Semi-Supervised Context Identification.} We formulate the context identification task as a semi-supervised learning problem, where we are given a labeled training set $\mathcal{D}^{aux}_l=\{(\x_i,c_i)\}_{i=1}^m$ and an unlabeled training set $\mathcal{D}^{aux}_u=\{\x_j\}_{j=1}^n$. Our target is to train a context identifier based on the labeled and unlabeled training examples, with the goal of boosting the performance of target multi-label alignment by exploiting contextual information. 

For every instance $\x$, according to \eref{eq:ap}, we obtain the global features $\g$. Compared to the local features, the global features capture the global information, which is beneficial for identifying the context. Then, by feeding $\g$ into a fully-connection layer and the softmax activation function, we obtain the predictions $\q^a$ (for weakly-augmented version) and $\p^a$ (for strongly-augmented version). For a unlabeled instances $\x_j$, we determine its pseudo-label $\hat{c}_j$ based on the probability $\p^a_j$ by $\hat{c}_j={\arg\max}_k p_{jk}^a$.
Then, we define the auxiliary loss, which comprises a supervised component and an unsupervised component, as follows:
\begin{equation}
	\begin{aligned}
		\mathcal{L}_{aux}=\sum_{i=1}^mL_{ce}(\p^a_i, c_i)+\sum_{j=1}^n\mathbf{1}(\max(\q^a_j)>\tau)L_{ce}(\p^a_j, \hat{c}_j),
	\end{aligned}
\end{equation}
where $\tau$ is the threshold to select reliable pseudo-labels \cite{sohn2020fixmatch}, and $L_{ce}$ is the cross entropy loss.
Finally, we define the overall loss function as:
\begin{equation}
	\mathcal{L}=\mathcal{L}_{sup}+\mathcal{L}_{unsup}+\mathcal{L}_{aux}.
\end{equation}

In general, the target and auxiliary tasks improve the performance of SSMLL from two perspectives: 1) by performing semantic-aware alignment, we enhance the recognition ability of the model via effective label-specific feature extraction; 2) by designing the context identification auxiliary task, we enhance the learning of feature representations by leveraging co-occurrence information. 

%% file: sec/4_experiment.tex
\vspace{-0.5em}
\section{Experiment}

\begin{table*}[!t]
	\centering

	\begin{tabular}{l c c c c| c c c c}
		\toprule
		\multirow{2}*{Method}
		& \multicolumn{4}{c}{VOC} & \multicolumn{4}{c}{COCO} \\
		\cmidrule(lr){2-9}
		& $p=0.05$ & $p=0.10$ & $p=0.15$ & $p=0.20$ & $p=0.05$ & $p=0.10$ & $p=0.15$ & $p=0.20$ \\ 
		\midrule
		BCE       & 67.95          & 75.35          & 78.19          & 79.38          & 58.90           & 63.75          & 65.91          & 67.33          \\
		ASL       & 71.46          & 78.00             & 79.69          & 80.77          & 59.12          & 63.82          & 66.10           & 67.51          \\
		\midrule
		Top-$1^*$      & 77.89          & 82.40         & 83.44         & 84.10          & 59.94          & 65.83          & 67.81          & 68.50          \\
		Top-$\text{k}^*$     & 76.60          & 80.44          & 82.26          & 83.30           & 62.39          & 66.97          & 68.24          & 68.66          \\
		$\text{IAT}^*$      & 75.00          & 77.63          & 81.67          & 82.77          & 62.93          & 66.18          & 67.46          & 68.56          \\
		\midrule
        CAP \cite{xie2023class}      & 76.16          & 82.16          & 83.48            & 84.41          & 62.43         &67.36          & 
        69.11     & 70.41               \\
		\midrule
		DualCoOp \cite{sun2022dualcoop}  & 79.16         & 83.69          & 84.55       &85.46             & 65.81          & 68.55          & 69.61            & 69.97      \\
        SCPNet \cite{ding2023exploring}    & 78.03            & 78.70        &79.68          &80.30              & 61.00          & 62.11         &
        62.49       &62.81              \\
        TaI \cite{guo2023texts}       & 79.14         & 82.93          & 84.14          & 84.64            & 67.02         & 69.31         & 69.94     
         & 70.25 \\
        DualCoOp++ \cite{hu2023dualcoop++} & 80.31  & 83.79          & 85.21             & 85.90
        & 67.19                 & 69.10         & 70.26             & 70.67 \\
		\midrule
		Ours      & \textbf{82.10} & \textbf{84.82} & \textbf{85.91} & \textbf{86.85} & \textbf{69.09} & \textbf{71.86} & \textbf{73.23} & \textbf{74.28} \\
		\bottomrule
	\end{tabular}
	
 \caption{\textbf{Comparison results on VOC and COCO in terms of mean Average Precision (mAP \%).} We compare our approach with SOTA methods under MLML problem setting and those methods based on fine-tuning CLIP. The best performance is highlighted in bold. Note that the symbol $^*$ denotes that combining DualCoOp \cite{sun2022dualcoop} with different pseudo labeling methods.}
 \label{tb:voc_coco}
\end{table*}

\begin{table*}[!t]
	\centering
	\small
	\begin{tabular}{l|ccccccccc|c}
		\toprule
		Method     & ASL  & Top-$1^*$  & Top-$\text{k}^*$  & $\text{IAT}^*$  & CAP    & DualCoOp  & SCPNet & TaI  & DualCoOp++  &Ours           \\
		\midrule
		$p = 0.05$ & 42.87 & 39.39 & 41.48 & 42.78  & 44.82 & 46.92 & 47.12     & 48.90     & 47.19 & \textbf{50.21} \\
		$p = 0.10$ & 46.50 & 44.79 & 45.89 & 45.83 & 48.24  & 48.82 & 48.89     & 50.77   & 48.81  & \textbf{51.65} \\
		$p = 0.15$ & 48.42 & 46.65 & 47.63 & 47.44  & 49.90 & 49.94 & 49.52     & 51.88  & 50.10  & \textbf{53.01}  \\
		$p = 0.20$ & 49.65 & 47.57 & 47.78 & 48.02  & 51.06 & 50.07 & 49.96     & 51.74  & 50.41  & \textbf{53.78} \\
		\bottomrule
	\end{tabular}
    \caption{\textbf{Comparison results on NUS in terms of mAP (\%).} The best performance is highlighted in bold. Note that the symbol $^*$ denotes that combining DualCoOp \cite{sun2022dualcoop} with different pseudo labeling methods.}
\label{tb:nus}
\end{table*}

\begin{table*}[!t]
\centering
\begin{tabular}{@{}ccc|cccccccc@{}}
\toprule
\multicolumn{3}{c|}{CBSA} & \multicolumn{4}{c}{COCO} & \multicolumn{4}{c}{VOC} \\ \midrule
TP & SAA & CI & \multicolumn{1}{l}{$p=0.05$} & \multicolumn{1}{l}{$p=0.10$} & \multicolumn{1}{l}{$p=0.15$} & \multicolumn{1}{l|}{$p=0.20$} & \multicolumn{1}{l}{$p=0.05$} & \multicolumn{1}{l}{$p=0.10$} & \multicolumn{1}{l}{$p=0.15$} & \multicolumn{1}{l}{$p=0.20$} \\ \midrule
\multicolumn{1}{l}{} & \multicolumn{1}{l}{} & \multicolumn{1}{l|}{} & 62.43 & 67.36 & 69.11 & \multicolumn{1}{c|}{70.41} & 76.16 & 82.16 & 83.48 & 84.41 \\ \midrule
\checkmark &  &  & 67.02 & 69.31 & 69.94 & \multicolumn{1}{c|}{70.25} & 79.14 & 82.93 & 84.14 & 84.64 \\
\checkmark & \checkmark &  & 68.65 & 71.39 & 72.96 & \multicolumn{1}{c|}{73.91} & 80.40 & 84.15 & 85.67 & 86.79 \\
\checkmark & \checkmark & \checkmark & \textbf{69.09} & \textbf{71.86} & \textbf{73.23} & \multicolumn{1}{c|}{\textbf{74.28}} & \textbf{82.10} & \textbf{84.82} & \textbf{85.91} & \textbf{86.85} \\ 
\bottomrule
\end{tabular}
\caption{\textbf{Ablation study.} Effects of different modules in our proposed method. We report the experiment results in terms of mAP (\%) on COCO and VOC datasets. The best performance is displayed in bold.}
\label{tb:ablation}
\end{table*}

\subsection{Experiment Settings}
\paragraph{Model Architecture.}  
To ensure a fair comparison with other methods, we employ ResNet-50 \cite{he2016deep} as the image encoder backbone of CLIP , and utilize the CLIP transformer as text encoder. Throughout training, the parameters of these two backbones remain frozen. We adopt the class-specific \cite{zhou2022learning} setting for prompts, where each category possesses an independent set of parameters. 
We optimize two independent context vectors with 16 context tokens following \cite{sun2022dualcoop}. To generate label-specific features, we use a transformer decoder consisting of 2 layers.
\vspace{-1em}
\paragraph{Datasets.} To assess the effectiveness of our method, we perform experiments on MS-COCO-2014 (COCO for short) \cite{lin2014microsoft}, VOC-2012 (VOC for short) \cite{everingham2015pascal}, and NUS-WIDE (NUS for short) \cite{chua2009nus}. Further details about these datasets can be found in the appendix due to space limit. We adopt label proportions within the range of $p\in\{0.05, 0.10, 0.15, 0.20\}$, and randomly sample from the dataset to construct the labeled training set, while leaving the rest without any annotations.
\vspace{-1em}
\paragraph{Methods.} To thoroughly validate the effectiveness of our method, we compare it with four groups of methods: 1) two MLL baseline methods that simply use BCE and ASL loss; 2) three methods from \cite{xie2023class} that utilize instance-aware pseudo-labeling techniques; 
3) one state-of-the-art SSMLL method CAP that uses class-wise thresholds to generate pseudo-labels; 4) four CLIP-based methods that fine-tune prompts to deal with partial labels. Note that for the last group of methods, we combine them with CAP to generate high-quality pseudo-labels.
\vspace{-1em}
\paragraph{Implementations Details.} We employ the clustering method to partition the original label space into $K=6$ clusters for MS-COCO, $K=2$ for VOC, and $K=2$ for NUS-WIDE. The optimization process utilizes AdamW optimizer and a one-cycle policy scheduler with a learning rate of 1e-3. For all datasets, we set the number of warm-up epochs to 8, and the total epochs to 40. The batch sizes are configured as 8, 64, and 64 for VOC, COCO, and NUS respectively. Additionally, the threshold for semi-supervised context identification is set to 0.9. All experiments are conducted on NVIDIA 3090 with a fixed seed of 1.

\begin{figure*}[!t]
  \centering
   \includegraphics[width=0.7\linewidth]{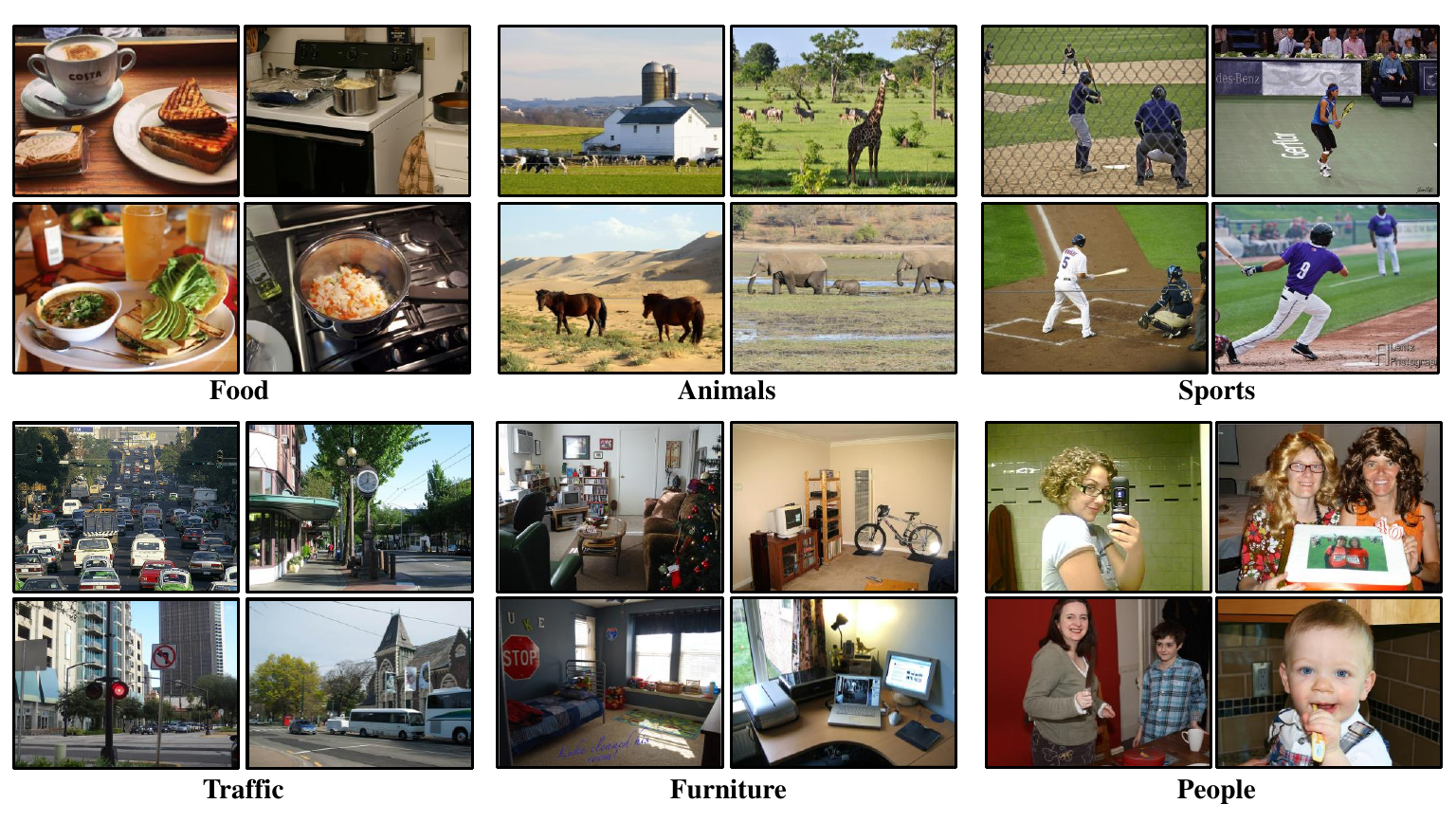}
   \caption{Visualization of images within various contexts that have top prediction probability.}
   \label{fig:context}
\end{figure*}
\vspace{-1em}
\subsection{Comparisons Results}

Experimental results on VOC, COCO and NUS are presented in \tref{tb:voc_coco} and \tref{tb:nus}. From the tables, we can see that: 1) abundant knowledge embedded in VLMs like CLIP could effectively alleviate the challenge of limited labeled data in SSMLL. 2) Although DualCoOp and TaI also adopt a dual prompts design and utilize the same thresholding strategy to obtain pseudo-labels, the performance of our method are significantly better than them. 
3) Our method achieves the best performance in all experimental settings and significantly outperforms other methods especially under small labeled ratios without introducing much higher computational cost. The training cost of different methods is presented in the appendix. The performance improvement can be attributed to the introduction of auxiliary task and better feature representation with our semantic-aware alignment target task. This validates that simplifying alignment task into a one-to-one problem, \textit{i.e.}, extracting label-specific image feature, is highly effective in dealing with multiple semantics in MLL. These experimental results demonstrate that our method exploits pre-training knowledge of CLIP more effectively.

\subsection{Ablation Study}
To comprehensively analyze the effect of different components in our method, we conducted an ablation study on COCO and VOC under various labeled ratio settings. 
As depicted in \tref{tb:ablation}, each module of our method demonstrates a positive effect on performance. In comparison to the baseline method CAP, fine-tuning CLIP by merely incorporating a target prompt (TP) exhibits a notable performance boost, particularly under small labeled ratio setting. On COCO dataset, TP leads to a 6.66\% mAP performance improvement when the ratio of labeled data $p=0.05$. 
Additionally, semantic-aware alignment (SAA) demonstrates a more significant performance increase as the labeled ratio increases, with the increase of 3.50\% and 2.15\% in mAP under $p=0.20$ on COCO and VOC respectively, which demonstrates that extracting label-specific feature from images leads to a better alignment between textual and visual modalities. With the introduction of the context identification auxiliary task, the performance further improves by an average of 0.67\%, which validates that the introduction of this auxiliary task could help model obtain accurate overall understanding of an image. 
To further explore the effect of auxiliary task context identification, we adjust the number of clusters in context partition task. As shown in \tref{tb:context_num}, our model achieves the best performance when context quantity $K=6$.

To validate the generalization ability of our method, we also conduct experiments on other settings of multi-label learning. Due to the page limit, we present the experiment results in \aref{sec:other_setting}.

\begin{figure}[!h]
	\centering
	\includegraphics[width=0.9\linewidth]{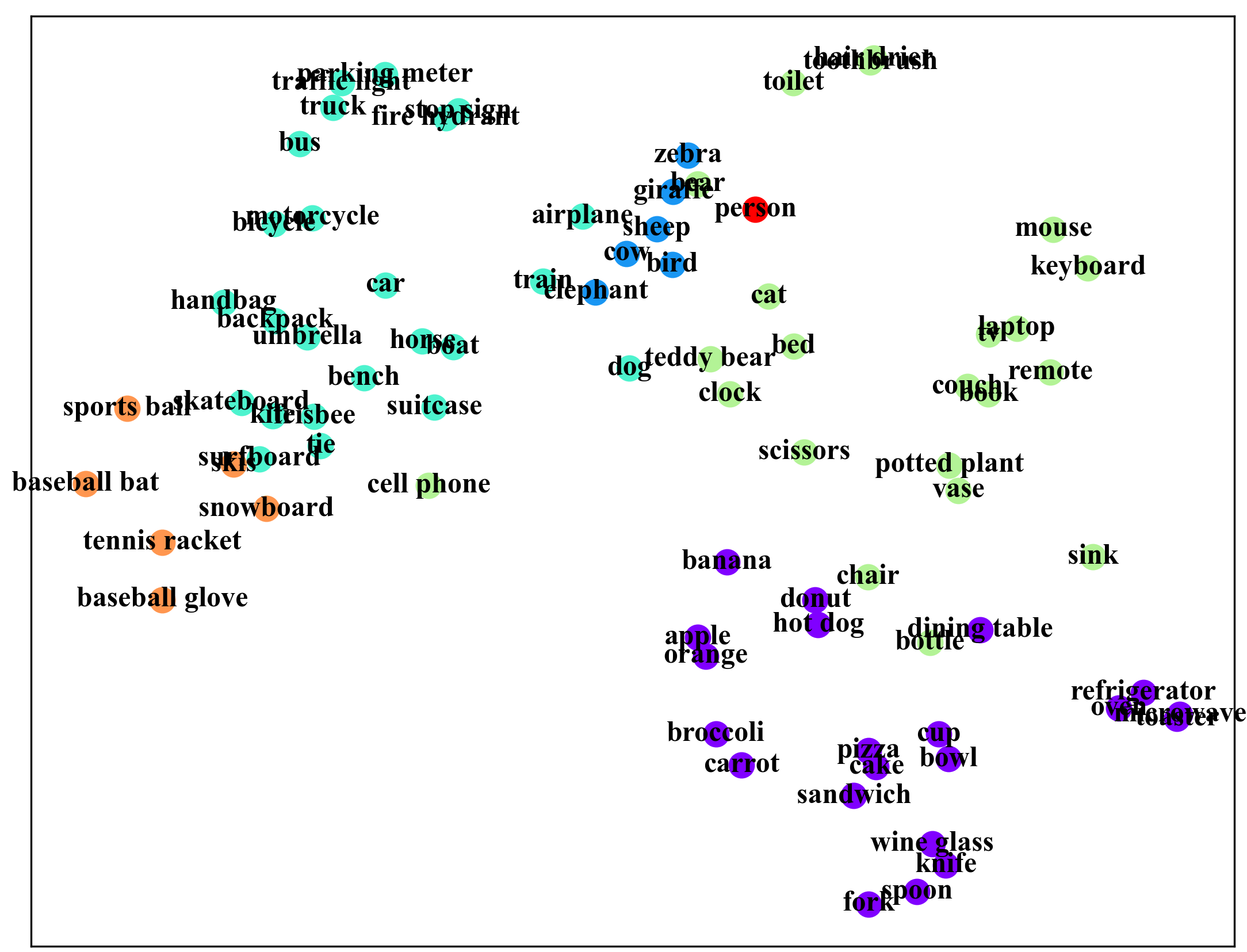}
	\caption{Context clusters generated by spectral clustering based on correlation matrix of COCO.}
	\label{fig:tsne}
\end{figure}

\begin{figure*}[!t]
\raggedright
  \begin{subfigure}{0.33\linewidth}
    \includegraphics[width=1.0\linewidth]{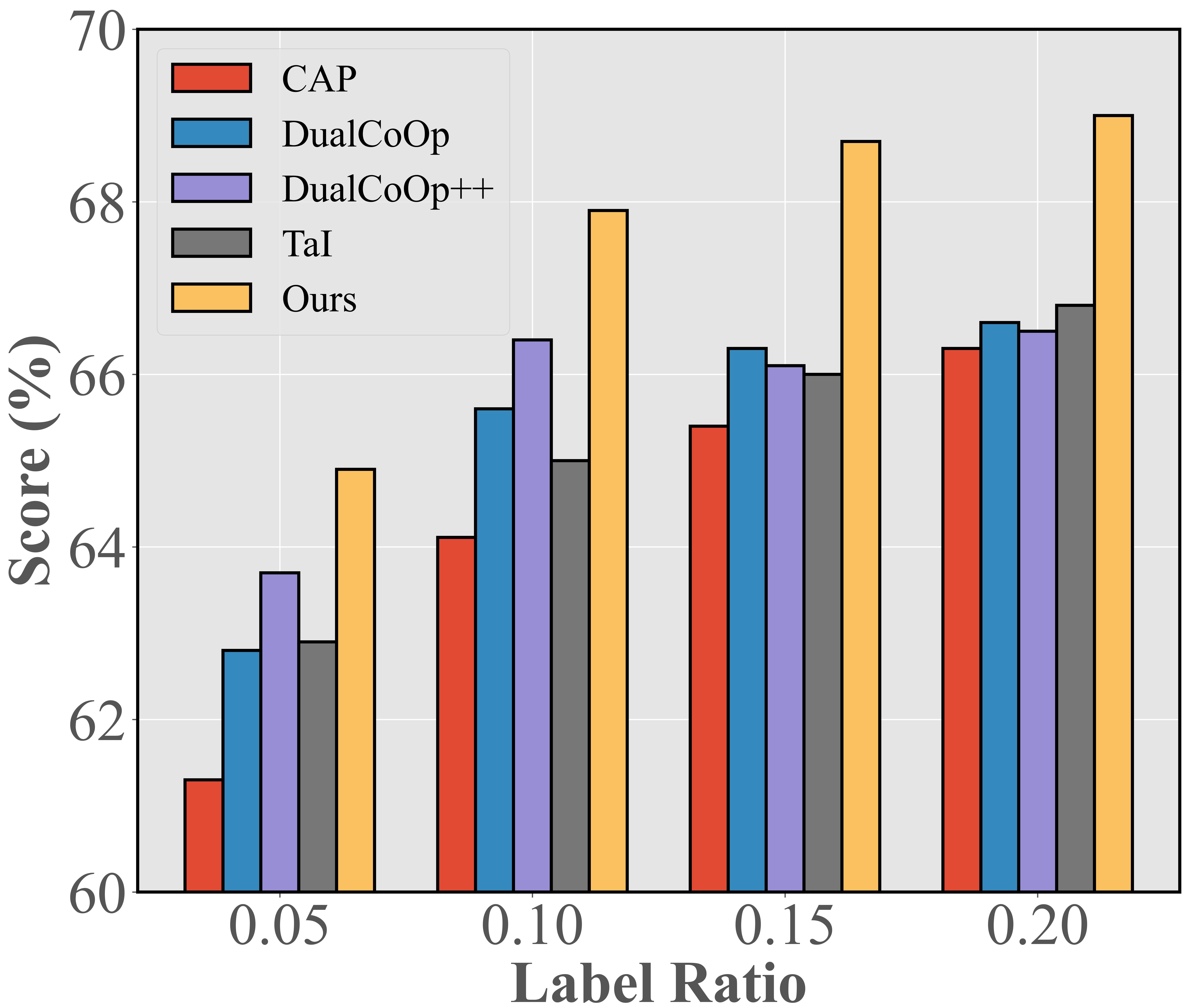}
    \caption{CF1 score on COCO}
    \label{fig:short-a}
  \end{subfigure}
    \centering
  \begin{subfigure}{0.33\linewidth}
    \includegraphics[width=1.0\linewidth]{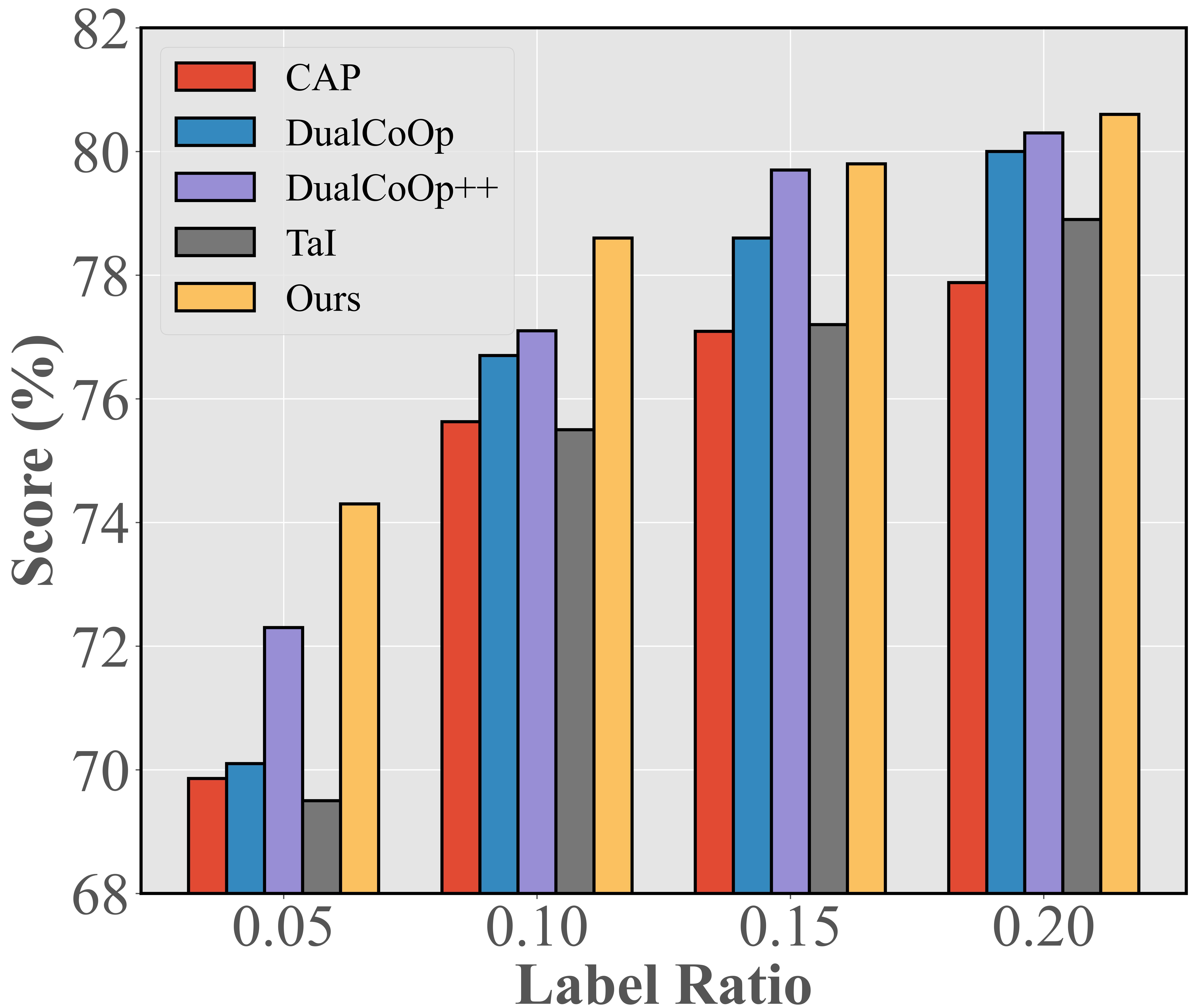}
    \caption{CF1 score on VOC}
    \label{fig:short-b}
  \end{subfigure}
  \raggedleft
   \begin{subfigure}{0.33\linewidth}
    \includegraphics[width=1.0\linewidth]{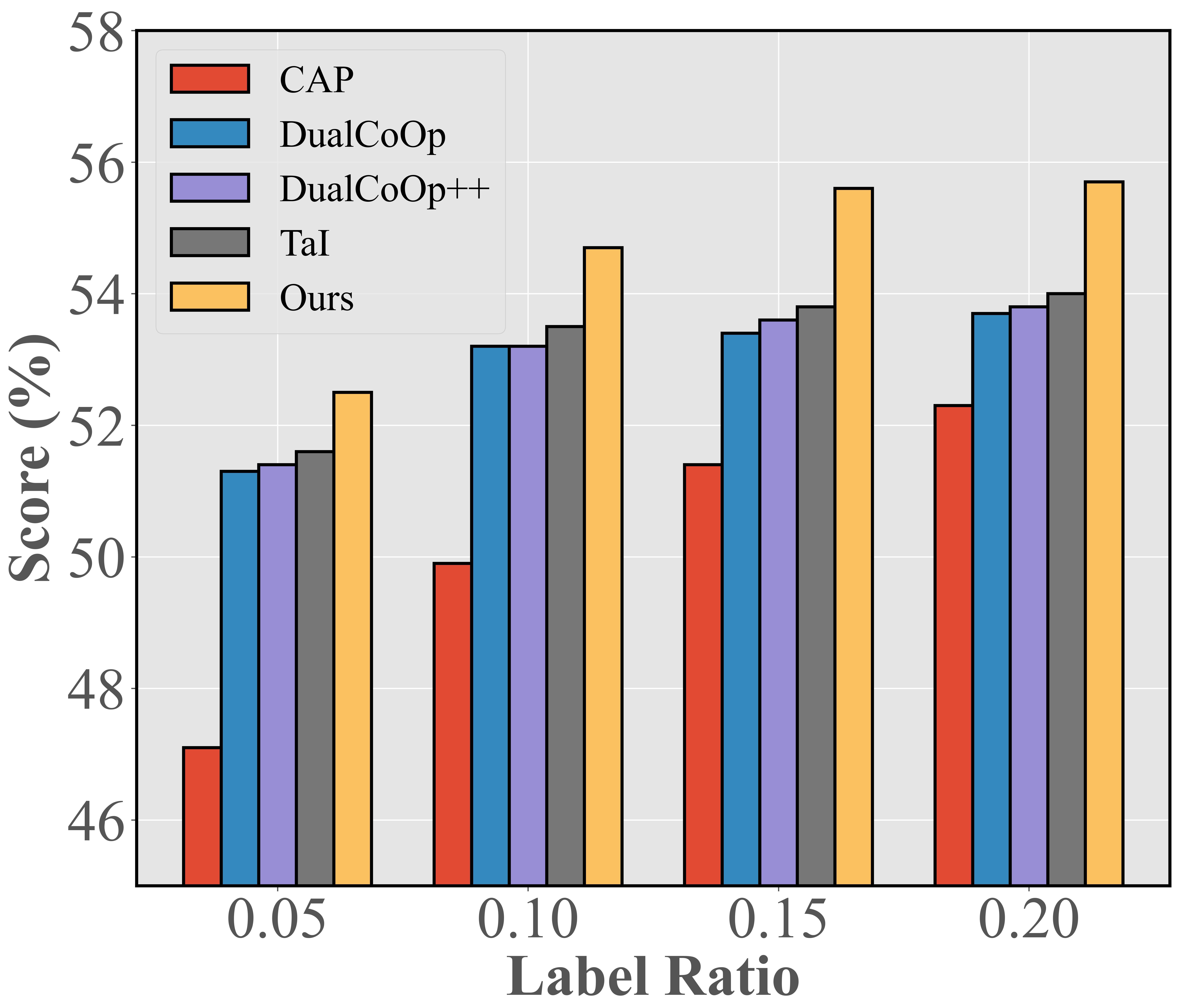}
    \caption{CF1 score on NUS}
    \label{fig:short-b}
  \end{subfigure}
  \caption{The quality of pseudo labels in terms of CF1 score on COCO, VOC and NUS.}
  \label{fig:CF1}
\end{figure*}

\begin{table}[t]
\centering
\begin{tabular}{@{}ccccc@{}}
\toprule
\multirow{2}{*}{\# Context} & \multicolumn{4}{c}{COCO} \\ \cmidrule(l){2-5} 
 & \multicolumn{1}{c}{$p = 0.05$} & \multicolumn{1}{c}{$p = 0.10$} & \multicolumn{1}{c}{$p = 0.15$} & \multicolumn{1}{c}{$p = 0.20$} \\ \midrule
$K=1$ & 68.65 & 71.39 & 72.96 & 73.91 \\
$K=4$ & 69.01 & 71.69 & 73.20 & 74.25 \\
$K=6$ & \textbf{69.09} & \textbf{71.86} & \textbf{73.23} & \textbf{74.28} \\ \bottomrule
\end{tabular}
\caption{The effect of different context quantity.}
\label{tb:context_num}
\end{table}

\subsection{Further Analysis}

In this section, we first present the clustering results of the COCO dataset using the correlation matrix of different categories. As depicted in \cref{fig:tsne}, with the context count $K$ set to 6 on COCO, the original label space is divided into 6 clusters. We can clearly observe that semantics that tend to co-occur are grouped into the same context. Based on the clustering results, we can partition the initial label space into 6 distinct subsets: \textit{food, animals, sports, traffic, furniture,} and \textit{people}.

\begin{figure}[!h]
	\centering
	\includegraphics[width=1.0\linewidth]{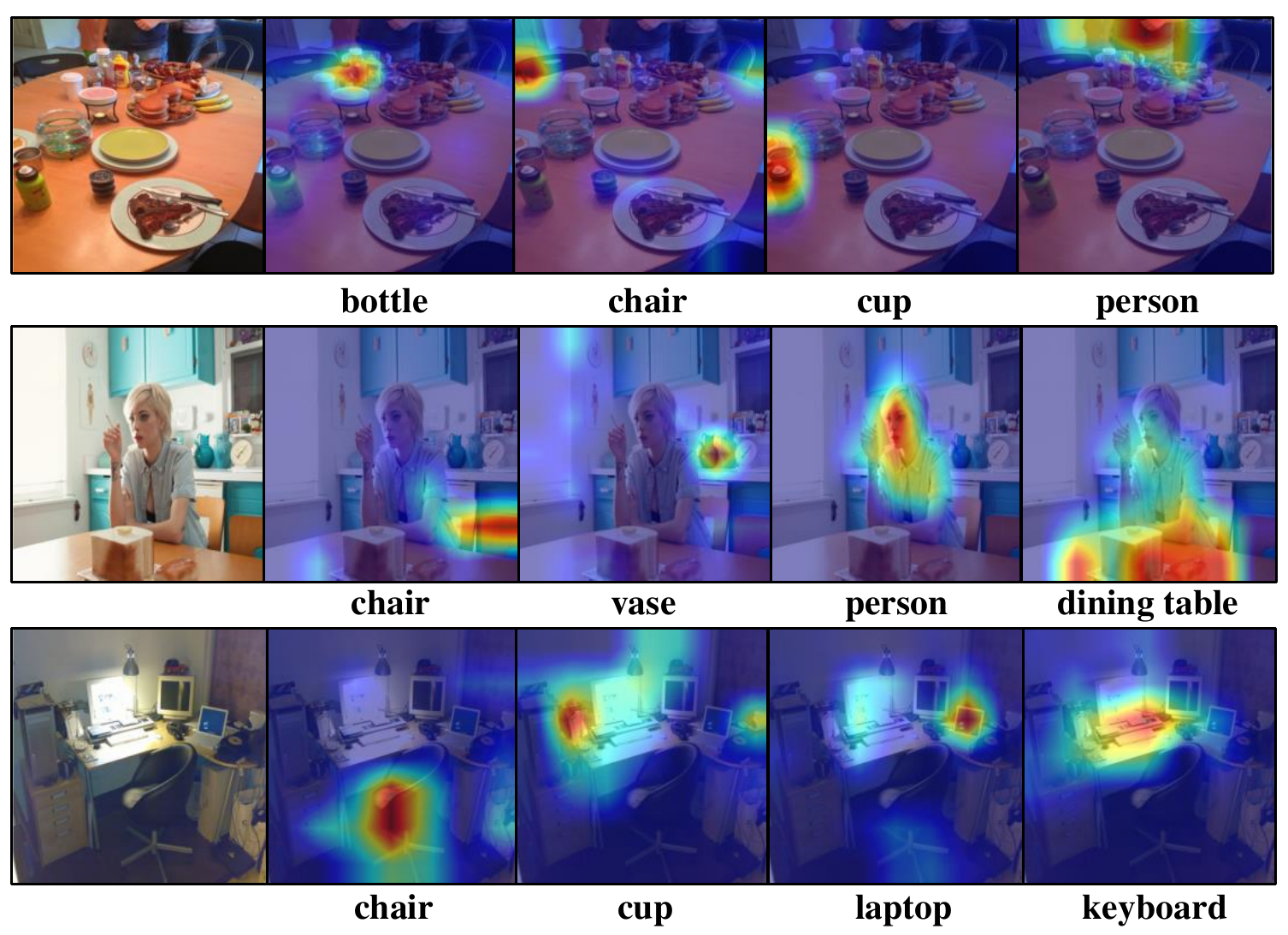}
	\caption{Visualization of different objects in COCO. Our model could effectively focus on the location of different semantics.}
	\label{fig:cam}
\end{figure}

For a direct visualization of the context identification results, we employ our trained context classifier to select several images within each context with high confidence scores, as demonstrated in \cref{fig:context}. From the result of context classification, we can see that the result is closely aligned with the clustering result based correlation matrix estimated on labeled dataset. For instance, within the context of \textit{food}, it is obvious that kitchens or tables are often present in the background of images belonging to this context, which means that with the introduction of context identification task, our method can effectively capture co-occurrence relationships, resulting in improved pseudo-labeling performance.


Furthermore, we evaluate the quality of pseudo labels in terms of CF1 score. \cref{fig:CF1} illustrates the result of CF1 scores on three datasets under different labeled ratios. We can clearly observe that our framework achieves the best performance under all labeled ratio settings, and achieves a more significant performance as the labeled ratio increases. This validates that our method could leverage supervised information more effectively and generate more accurate pseudo labels to boost the learning on unlabeled dataset. 

In order to further demonstrate the model's capability to locate various semantic information, we adopt CAM \cite{selvaraju2017grad} technique to visualize where the model is placing its attention. As shown in \cref{fig:cam}, we can see that our model has a powerful ability to locate different semantics, including challenging targets like tiny objects, such as \textit{cup} and \textit{vase}. These results demonstrate that our method is able to significantly enhance the recognition ability of the model based on limited labeled data, resulting in favorable pseudo-labeling performance. 

\vspace{-0.5em}
\section{Conclusion}
In this paper, we propose a novel framework to tackle the problem of semi-supervised multi-label learning, which is the first work that adapts vision-language model, \textit{i.e.}, CLIP, under this problem setting. With the target of achieving better alignment between textual and visual modalities, we equip our model with semantic-aware prompts to extract label-specific image features. To incorporate the model with comprehensive understanding of image, we design a new auxiliary task, context identification, which benefits the main classification task by exploiting co-occurrence relationships. Furthermore, we formulate context identification in a semi-supervised learning manner to extract more supervised information from limited labeled data, benefiting the learning process on unlabeled samples by enhancing the quality of pseudo-labels. Extensive experiments on multiple benchmarks have demonstrated that our method can achieve state-of-the-art performance. In the future, we will further study how to generalize our method into other problem settings, and align textual and visual modalities more closely to further increase model performance.

%% file: sec/X_suppl.tex
\clearpage
\setcounter{page}{1}
\maketitlesupplementary

\begin{table*}
\centering
\begin{tabular}{@{}c|c|cccccccc@{}}
\toprule
Dataset               & Methods  & 10\%          & 20\%          & 30\% & 40\% & 50\% & 60\% & 70\% & 80\% \\ \midrule
\multirow{9}{*}{COCO} & SSGRL \cite{chen2019learning} & 62.5          & 70.5          & 73.2 & 74.5 & 76.3 & 76.5 & 77.1 & 77.9 \\
                      & GCN-ML \cite{chen2019multi}  & 63.8          & 70.9          & 72.8 & 74.0 & 76.7 & 77.1 & 77.3 & 78.3 \\
                      & SST \cite{Chen2022SST}     & 68.1          & 73.5          & 75.9 & 77.3 & 78.1 & 78.9 & 79.2 & 79.6 \\
                      & SARB \cite{pu2022semantic}    & 71.2          & 75.0          & 77.1 & 78.3 & 78.9 & 79.6 & 79.8 & 80.5 \\
                      & HST \cite{Chen2024HST}     & 70.6          & 75.8          & 77.3 & 78.3 & 79.0 & 79.4 & 79.9 & 80.2 \\
                      & DSRB \cite{pu2024dual}    & 72.5          & 76.0          & 77.6 & 78.7 & 79.6 & 79.8 & 80.0 & 80.5 \\ 
                      & DualCoOp \cite{sun2022dualcoop} & 78.7          & 80.9          & 81.7 & 82.0 & 82.5 & 82.7 & 82.8 & 83.0 \\
                      & SCPNet \cite{ding2023exploring}  & \textbf{80.3} & \textbf{82.2} & 82.8 & 83.4 & 83.8 & 83.9 & 84.0 & 84.1 \\
 & Ours (SAA) & 79.8 & 81.9 & \textbf{83.5} & \textbf{84.3} & \textbf{84.4} & \textbf{84.7} & \textbf{85.0} & \textbf{85.3} \\ \bottomrule
\end{tabular}
\caption{Comparison with methods in partial multi-label setting on COCO.}
\label{tab:other_setting}
\end{table*}




\section{Experiment Settings}
\label{sec:experiment setting}
\subsection{Details of Datasets}

We conduct experiments on three benchmark datasets: MS-COCO-2014, VOC-2012, and NUS-WIDE. MS-COCO comprises a training set with 82,081 images and a test set with 40,137 images across 80 different classes. VOC-2012 consists of 5,717 images for training and 5,823 images for testing distributed among 20 classes. NUS-WIDE contains 150,000 images across 81 classes, featuring various resolutions for training, and 60,200 images for testing. These three datasets are widely used in SSMLL field to evaluate model performance. Compared with VOC, COCO and NUS have more semantics per image, which means that it's more challenging for model to correctly identify all objects in images.

\subsection{Training Details}
To ensure a fair comparison, we employ identical augmentation techniques across all methods, following the approach in \cite{xie2023class}. Specifically, we apply both weak and strong augmentations to every image. Initially, each image is resized to $224 \times 224$. For the weakly-augmented version, a random horizontal flip is performed. We adopt RandAugment and Cutout for the strongly-augmented version of image. 
We employ random initialization for learnable prompts in all CLIP-based methods and design class-specific prompts (CSP) following \cite{zhou2022learning}. In this design, each class is associated with its own set of context tokens. 
\section{Additional Experiment Results}
\label{sec:more results}

\subsection{Performance on Partial Multi-label Setting}
\label{sec:other_setting}
We conduct additional experiments on COCO in partial multi-label setting to further validate the effectiveness of our method, using the same experimental setup as in DualCoOP \cite{sun2022dualcoop} and SCPNet \cite{ding2023exploring} with ResNet-101 as the visual encoder of CLIP and all the images are resized into 448$\times$448. To create training samples with partial labels, we randomly mask out labels from complete annotation, and the proportion of available labels ranges from 10\% to 80\%. 

In the partial multi-label setting, complete label annotations are unavailable, preventing us from obtaining an accurate co-occurrence matrix to leverage label correlations among categories and enhance the overall understanding of images. As a result, only the semantic-aware alignment module is used. We compare our method against two groups of baselines: (1) six conventional partial multi-label learning methods, including SSGRL \cite{chen2019learning}, GCN-ML \cite{chen2019multi}, SST \cite{Chen2022SST}, SARB \cite{pu2022semantic}, HST \cite{Chen2024HST}, and DSRB \cite{pu2024dual}; (2) two methods based on fine-tuning learnable prompts in CLIP, namely, DualCoOP \cite{sun2022dualcoop} and SCPNet \cite{ding2023exploring}. As shown in \tref{tab:other_setting}, our method demonstrates improved performance as the proportion of available labels increases, indicating that extracting label-specific image features can achieve a more compact alignment between the two modalities.

\begin{table}[H]
\centering
\begin{tabular}{@{}c|cccc@{}}
\toprule
\multirow{2}{*}{Initialization} & \multicolumn{4}{c}{COCO}                                          \\ \cmidrule(l){2-5} 
                                & 0.05           & 0.10            & 0.15           & 0.2            \\ \midrule
Random                          & \textbf{69.09} & \textbf{71.86} & \textbf{73.23} & 74.28          \\
Template                        & 68.89          & 71.85          & 73.22          & \textbf{74.34} \\ \bottomrule
\end{tabular}
\caption{Comparison of different initialization techniques. The best performance is displayed in bold (\%).}
\label{tb:template}
\end{table}

\subsection{Prompt Settings}
\paragraph{Comparison of Initialization Techniques.} 
We analyze the effect of different initialization techniques for prompts. Here, we compare the results of random initialization vs template initialization. For the template initialization, we construct a simple prompt template, \textit{A photo of a [CLS]}, where [CLS] is replaced with actual class names. As depicted in \tref{tb:template}, we observe similar performance with both initialization techniques. Thus different initialization methods have little effect on model's performance. So we simply use random initialization technique in all experiments.

\begin{table}[H]
\centering
\begin{tabular}{@{}c|cccc@{}}
\toprule
\multirow{2}{*}{\begin{tabular}[c]{@{}c@{}}Prompt Length\end{tabular}} & \multicolumn{4}{c}{COCO}      \\ \cmidrule(l){2-5} 
& 0.05  & 0.10   & 0.15  & 0.20   \\ \midrule
8  & \textbf{69.17} & 71.76 & 73.18 & 74.41 \\
16 & 69.09 & 71.86 & 73.23 & 74.28 \\
32 & 68.91 & 71.75 & \textbf{73.27} & 74.26 \\
48 & 68.94 & \textbf{71.92} & 73.14 & \textbf{74.34} \\
64 & 69.09 & 71.91 & 73.26 & 74.24 \\ \bottomrule
\end{tabular}
\caption{The effect of prompt length. The best performance is displayed in bold (\%).}
\label{tb:context length}
\end{table}

\paragraph{Prompt Length.} 
To investigate the effect of prompt length, we perform experiments with varying prompt length settings on the COCO dataset. The experimental results are presented in \tref{tb:context length}. Notably, the results demonstrate that different prompt lengths result in similar performance. This demonstrates that our method can achieve promising results without introducing high computational cost. In this case, we simply follow the same prompt length setting as in \cite{sun2022dualcoop} to ensure the fairness.

\paragraph{Single vs Class-Specific Prompts.} 
A single learnable prompt optimizes the same set of parameters across all classes, whereas the class-specific prompt (CSP) setting uses an individual set of optimizable parameters for each category. To discover the influence of prompt setting, we present the comparison result for these two approaches. As shown in \tref{tb:csc}, CSP consistently outperforms the single-prompt approach across all labeled settings on the COCO dataset, achieving a maximum improvement of 0.34\%. This improvement demonstrates that optimizing the alignment between label-specific image features and class-specific text features effectively simplifies the original problem. Clearly, using independent prompts for each class enables a more effective alignment between the visual and textual modalities.

\begin{table}
\centering
\begin{tabular}{@{}c|cccc@{}}
\toprule
\multirow{2}{*}{\begin{tabular}[c]{@{}c@{}}Prompt Setting\end{tabular}} & \multicolumn{4}{c}{COCO}                                          \\ \cmidrule(l){2-5} 
 & 0.05           & 0.10            & 0.15           & 0.20            \\ \midrule
CSP  & \textbf{69.09} & \textbf{71.86} & \textbf{73.23} & \textbf{74.28} \\
w/o CSP  & 68.93          & 71.52          & 72.96          & 74.13          \\ \bottomrule
\end{tabular}
\caption{Comparison of different prompt setting. The best performance is displayed in bold (\%).}
\label{tb:csc}
\end{table}


\begin{table}
\centering
\resizebox{0.47\textwidth}{!}{
\begin{tabular}{@{}cccc@{}}
    \toprule
    Methods    & \begin{tabular}[c]{@{}c@{}}Training Latency\\ (ms/img) $\downarrow$\end{tabular} & \begin{tabular}[c]{@{}c@{}}Training Memory\\ (GB/img) $\downarrow$ \end{tabular} & \begin{tabular}[c]{@{}c@{}}mAP\\ (\%) $\uparrow$\end{tabular} \\ \midrule
    CAP  & 2.436 & 0.114  & 62.43  \\ \midrule
    DualCoOp++ & 3.392 & 0.107  & 67.17  \\
    TaI   & 3.046  & 0.108  & 67.02  \\
    SCPNet   & 3.257  & 0.113  & 61.00  \\
    DualCoop  & 3.314  & 0.107  & 65.81  \\ \midrule
    Ours  & 3.533  & 0.120  & 69.09  \\ \bottomrule
\end{tabular}
}
 \caption{Computational cost under $p=0.05$ on COCO.}
 \label{tab:usage}
\end{table}

\subsection{Computational Cost}
\label{sec:cost}
We conduct additional experiments to compare the training cost of our method against baseline models. These experiments were performed on the COCO dataset under a $p=0.05$ setting to evaluate the training efficiency of various approaches. As shown in \tref{tab:usage}, our method incurs only a slight increase in training latency and VRAM usage while achieving significant performance gains under the $p=0.05$ setting on COCO, which demonstrates that the introduction of semantic-aware alignment module and context identification task could achieve more compact alignment between visual and textual modalities and fully utilize contextual information to enhance feature representations.

\subsection{Error Bars}
In this section, we provide the error bar report on COCO dataset under different label ratios. We conduct 3 independent experiments with fixed random seeds 1, 2 and 3 respectively. And we choose 95\% confidence interval and report standard error of the mean. As depicted in \tref{tab:error}, our method demonstrates steady performance under all label ratio settings.

\begin{table}[H]
\centering
\resizebox{0.47\textwidth}{!}{
\begin{tabular}{@{}ccccc@{}}
    \toprule
    Methods & $p = 0.05$ & $p = 0.10$ & $p = 0.15$ & $p=0.20$ \\ 
    \midrule
    Ours    & 69.05$\pm$0.064 & 71.77$\pm$0.146 & 73.15$\pm$0.357 & 74.15$\pm$0.250       \\ \bottomrule
\end{tabular}
}
  \caption{Error bar report on COCO.}
	\label{tab:error}
\end{table}